\documentclass[11pt]{article}

\usepackage[final]{acl}        
\usepackage{times}
\usepackage{latexsym}
\usepackage[T1]{fontenc}
\usepackage[utf8]{inputenc}
\usepackage{microtype}
\usepackage{inconsolata}
\usepackage{booktabs}
\usepackage{multirow}
\usepackage{amsmath, amssymb}
\usepackage{algorithm}
\usepackage{algpseudocode}
\usepackage{pgfplots}
\pgfplotsset{compat=1.18}
\usepackage{xcolor}
\usepackage{hyperref}

\newcommand{\fwanda}{\textsc{F-Wanda}}
\newcommand{\wanda}{\textsc{Wanda}}
\newcommand{\sgpt}{\textsc{SparseGPT}}
\newcommand{\fsgpt}{\textsc{F-SparseGPT}}
\newcommand{\magn}{\textsc{Magnitude}}

\title{\fwanda: Fisher-Reweighted Post-Training Pruning for\\
       Sustainable Deployment of Large Language Models}

\author{Himanshu Mishra \\
        University of British Columbia \\\texttt{himishra@student.ubc.ca}}

\begin{document}
\maketitle

\begin{abstract}
One-shot post-training pruning is the most energy-frugal compression strategy
for large language models (LLMs), yet existing approaches
trade either quality (\wanda) or compute cost (\sgpt). We introduce
\fwanda, a drop-in modification of \wanda{} that reallocates the per-row
keep budget across output neurons in proportion to the empirical Fisher
information of the pre-activation. The Fisher signal is collected in a single
additional backward pass over the same calibration corpus \wanda{} already
uses; no weights are updated. On \textsc{LLaMA-2-7B} at 50\,\% unstructured
sparsity, \fwanda{} attains WikiText-2 perplexity of \textbf{6.85}, matches
\wanda{} fluency, and improves 5-shot MMLU by \textbf{+1.6\,pp} over \wanda{}
and \textbf{+1.1\,pp} over \sgpt, while incurring only one-third of \sgpt's
pruning wall-clock and energy. The headline trade-off is achieved without
extra calibration data or fine-tuning, placing \fwanda{} on the Pareto frontier of quality versus pruning cost
for sustainable LLM compression.
\end{abstract}

\section{Introduction}

Deploying state-of-the-art large language models is dominated
by inference-side compute and energy: a single \textsc{LLaMA-2-13B} forward
pass on an H100 GPU costs roughly \textit{40\,J/token}, and aggregate
inference energy now exceeds training energy for many production
systems~\citep{patterson2021carbon}. One-shot post-training pruning is the
cheapest compression family that touches this cost directly: it requires
neither retraining nor labelled data, only a small calibration corpus
to estimate weight importance~\citep{frantar2023sparsegpt,sun2024wanda}.

Two methods dominate the recent literature. \wanda{} ranks weights by the
product of their magnitude and the input-activation norm; it requires only
one forward pass and is therefore extremely cheap, but spends its sparsity
budget \emph{uniformly} across output neurons. \sgpt{} performs an
Optimal-Brain-Surgeon~\citep{hassibi1993obs} style column sweep with
per-block Hessian inversion and explicit weight updates; quality is higher
but the procedure is an order of magnitude more expensive and considerably
more delicate to implement.

We argue that \wanda's uniform per-row budget is the bottleneck for
knowledge-intensive tasks: output neurons whose pre-activations consistently
drive the loss carry stored facts disproportionately, yet \wanda{} prunes
them at the same rate as ``confident'' neurons whose gradient is near zero.
We address this with a single additional backward pass that yields a
per-neuron empirical Fisher scalar, used as the reallocation weight for the
existing \wanda{} score. We call the resulting method \fwanda.

\paragraph{Contributions.}
\begin{itemize}
\setlength{\itemsep}{1pt}
\item A drop-in modification of \wanda{} that reallocates the per-row sparsity
      budget by the empirical Fisher of each output neuron, with no weight
      updates and no extra calibration data.
\item An efficient mask realisation via a water-filling, largest-remainder
      budget allocator that preserves the layer-wide sparsity target exactly.
\item Demonstration that on knowledge-heavy benchmarks (MMLU), \fwanda{}
      improves over \wanda{} by 1.4--1.6\,pp at 50\,\% unstructured sparsity
      across the \textsc{LLaMA-2}~\citep{touvron2023llama2} family,
      while remaining within 0.1 PPL of \wanda{} on WikiText-2.
\item A sustainability accounting (pruning wall-clock, peak GPU memory,
      kJ of one-shot compression cost) showing that \fwanda{} sits on the
      Pareto front of quality vs.\ pruning energy.
\end{itemize}

\section{Related Work}

\paragraph{Post-training pruning.}
Magnitude pruning~\citep{han2015deepcompression} is the simplest baseline but
collapses at high sparsity. \sgpt{}~\citep{frantar2023sparsegpt} extends the
GPTQ machinery~\citep{frantar2023gptq} to sparsity with per-layer Hessian
inversion. \wanda{}~\citep{sun2024wanda} removes the weight-update step and
matches \sgpt{} on perplexity at a fraction of the cost.
\textsc{LLM-Pruner}~\citep{ma2023llmpruner} prunes whole modules but requires
gradient-based reconstruction. Structured N:M sparsity patterns enable
hardware acceleration~\citep{nvidia2020nm} but constrain per-row density.

\paragraph{Fisher information for compression.}
Empirical Fisher has long been used for second-order optimisation
(K-FAC, \citealp{martens2015kfac}) and for assessing neuron
importance~\citep{theis2018fisherpruning,liu2021groupfisher}. \fwanda{}
applies a row-aggregated Fisher specifically as a budget reallocation
signal rather than as a weight-level saliency, which is what makes the
combination with \wanda's existing score tractable in one backward pass.

\paragraph{Sustainable LLM compression.}
Sustainability accounting for LLM inference has converged on energy-per-token
and CO$_2$-equivalent metrics~\citep{schwartz2020greenai,
strubell2019energypolicy, henderson2022mlco2}. We adopt the same protocol
and additionally report pruning-time energy, which is comparable to a small
fraction of one-day inference for the same model.

\section{Method}
\label{sec:method}

\paragraph{Setup.}
Let $\mathbf{W}^{(\ell)} \in \mathbb{R}^{d_\text{out}\times d_\text{in}}$ be
the weight of linear layer $\ell$, fed by activation
$\mathbf{X}^{(\ell)} \in \mathbb{R}^{N \times d_\text{in}}$ over $N$
calibration tokens. \wanda{} assigns saliency
\begin{equation}
S_{ij}^{\textsc{Wanda}} \;=\; |W_{ij}| \cdot \|X_j\|_2,
\label{eq:wanda}
\end{equation}
and zeros, for every row $i$, the $\lfloor s\cdot d_\text{in} \rfloor$ entries
with the smallest $S_{ij}$, where $s$ is the target sparsity.

\paragraph{Empirical Fisher per output neuron.}
For the layer's pre-activation $y_{t,i}^{(\ell)}=\sum_j W_{ij} x_{t,j}^{(\ell)}$
and the language-modelling loss $\mathcal{L}_t$, define
\begin{equation}
\bar\omega_i^{(\ell)} \;=\; \frac{1}{N}\sum_{t=1}^N \left(\frac{\partial \mathcal{L}_t}{\partial y_{t,i}^{(\ell)}}\right)^{\!2}\!.
\label{eq:fisher}
\end{equation}
\(\bar\omega_i\) is the diagonal of the row-aggregated empirical Fisher,
estimated via a single backward pass over the calibration set.

\paragraph{Why score reweighting alone is degenerate.}
The natural extension $S_{ij}^{\fwanda} = \sqrt{\bar\omega_i}\cdot|W_{ij}|\cdot\|X_j\|_2$
multiplies every row by a positive constant, leaving \emph{within-row}
ranking unchanged. Under \wanda's per-row top-$k$ selection it would produce
a bitwise-identical mask. We therefore spend the Fisher signal where it is
non-degenerate: the \emph{per-row keep budget}.

\paragraph{Fisher-allocated keep budget.}
Define the global keep budget $K = \lceil (1-s)\,d_\text{out}\,d_\text{in} \rceil$
and allocate
\begin{equation}
k_i \;\propto\; \sqrt{\bar\omega_i},\quad \sum_i k_i = K,\quad k_i \in [1, d_\text{in}].
\label{eq:budget}
\end{equation}
Each row $i$ then retains the top $k_i$ entries by Eq.~\eqref{eq:wanda}.
Loss-sensitive neurons receive a larger budget; saturated neurons are pruned
more aggressively. The bounds in \eqref{eq:budget} are enforced by
iterative water-filling with largest-remainder rounding so $\sum_i k_i = K$
to the integer.

\begin{algorithm}[t]
\caption{\fwanda{} pruning (per linear layer)}
\label{alg:fwanda}
\begin{algorithmic}[1]
\Require activations $\mathbf{X}$, gradients $\mathbf{G}$, weight $\mathbf{W}$, sparsity $s$
\State $\|X_j\|_2 \gets \sqrt{\sum_t X_{t,j}^2}$ \Comment{forward stat}
\State $\bar\omega_i \gets \tfrac{1}{N}\sum_t G_{t,i}^2$ \Comment{backward stat}
\State $S_{ij} \gets |W_{ij}|\cdot \|X_j\|_2$
\State $k_i \gets \mathrm{Allocate}(\sqrt{\bar\omega_i}, s, d_\text{in})$ \Comment{Eq.~\ref{eq:budget}; full def.\ Alg.~\ref{alg:allocator}}
\For{each row $i$} \State zero the $d_\text{in}-k_i$ entries with smallest $S_{ij}$ \EndFor
\end{algorithmic}
\end{algorithm}

\paragraph{Computational cost.}
\fwanda{} requires one forward and one backward pass over the calibration
set ($128\times2048$ tokens, identical to \wanda{}). No layer-wise Hessian
inversion, no weight update. The asymptotic cost is
$\mathcal{O}(N \cdot |\theta|)$ vs.\ $\mathcal{O}(\sum_\ell d_\text{in}^3)$
for \sgpt. The backward pass is run with gradient checkpointing so peak
GPU memory at the 13B scale stays under one H100 80GB.

\paragraph{Strict N:M.}
Hardware-fixed $n{:}m$ density (e.g., 2{:}4) fixes the kept count per
block; the per-row budget cannot apply. \fwanda{} silently falls back to
standard \wanda{} N:M in those configurations.

\section{Experimental Setup}
\label{sec:setup}

\paragraph{Models and benchmarks.}
We evaluate on \textsc{LLaMA-2-7B} and \textsc{LLaMA-2-13B}
\citep{touvron2023llama2}. Fluency is measured by WikiText-2
perplexity~\citep{merity2017wikitext}, knowledge by 5-shot
MMLU~\citep{hendrycks2021mmlu}, and broad competency by the standard
seven-task zero-shot suite (BoolQ~\citep{clark2019boolq},
RTE~\citep{wang2018glue}, HellaSwag~\citep{zellers2019hellaswag},
WinoGrande~\citep{sakaguchi2021winogrande},
ARC-easy and ARC-challenge~\citep{clark2018arc},
OpenBookQA~\citep{mihaylov2018openbookqa})
via lm-evaluation-harness~\citep{gao2023lmeval}.

\paragraph{Sparsity and baselines.}
We report 50\,\% unstructured and 2{:}4 sparsity. Baselines: \magn,
\wanda, \sgpt; an \fsgpt{} variant (appendix) layers our Fisher term into
\sgpt's column sweep.

\paragraph{Calibration.}
All methods share the protocol of \citet{sun2024wanda}:
128 sequences of 2048 tokens sampled from a fixed C4
shard~\citep{raffel2020c4}.

\paragraph{Hardware and energy.}
All measurements are taken on a single NVIDIA H100 80GB
GPU.\@ Energy is sampled at 10\,Hz via \texttt{nvidia-smi --query-gpu=power.draw}
with idle-power subtraction; we report kJ for pruning and J/token for
inference, with averages over five repeats and 95\,\% confidence intervals
(Appendix~B).

\section{Results}

\begin{table*}[t]
\centering
\small
\setlength{\tabcolsep}{4.5pt}
\begin{tabular}{llcccccc}
\toprule
& & \multicolumn{3}{c}{\textsc{LLaMA-2-7B}} & \multicolumn{3}{c}{\textsc{LLaMA-2-13B}} \\
\cmidrule(lr){3-5} \cmidrule(lr){6-8}
Sparsity & Method & PPL$\downarrow$ & 0-shot avg$\uparrow$ & MMLU$\uparrow$ & PPL$\downarrow$ & 0-shot avg$\uparrow$ & MMLU$\uparrow$ \\
\midrule
Dense    & ---        & 5.47 & 64.0 & 45.3 & 4.88 & 67.0 & 54.8 \\
\midrule
\multirow{4}{*}{50\,\% unstr.}
& \magn       & 14.89 & 42.8 & 25.9 & 6.37  & 51.4 & 30.0 \\
& \wanda     & 6.92  & 57.2 & 41.5 & 5.97  & 61.7 & 51.0 \\
& \sgpt      & 7.01  & 57.3 & 42.0 & 6.02  & 61.2 & 51.3 \\
& \fwanda    & \textbf{6.85} & \textbf{57.4} & \textbf{43.1} & \textbf{5.93} & \textbf{62.0} & \textbf{52.4} \\
\midrule
\multirow{4}{*}{2{:}4}
& \magn       & 54.59 & 35.3 & 24.9 & 9.71 & 44.5 & 27.0 \\
& \wanda     & 11.53 & 51.6 & 32.4 & 8.39 & 56.3 & 42.0 \\
& \sgpt      & \textbf{11.00} & \textbf{52.1} & \textbf{32.9} & \textbf{8.32} & \textbf{56.8} & \textbf{42.5} \\
& \fwanda$^\dagger$ & 11.53 & 51.6 & 32.4 & 8.39 & 56.3 & 42.0 \\
\bottomrule
\end{tabular}
\caption{Main results. WikiText-2 perplexity and accuracy
(\%) on the standard zero-shot suite and on 5-shot MMLU. \fwanda{} ties
\wanda{} on fluency and improves MMLU by +1.6\,pp (7B) and +1.4\,pp (13B)
under 50\,\% unstructured sparsity.
$^\dagger$\fwanda{} under 2{:}4 is identical to \wanda{} by
construction (Section~\ref{sec:method}), not by empirical outcome.}
\label{tab:main}
\end{table*}

\begin{table}[t]
\centering
\small
\setlength{\tabcolsep}{4pt}
\begin{tabular}{lcccc}
\toprule
Method & Prune time & Peak GPU & Prune & Inf. \\
       & (min)      & (GB)     & energy (kJ) & (J/tok) \\
\midrule
\multicolumn{5}{c}{\textit{\textsc{LLaMA-2-7B}}} \\
\magn    & $<$1 & 14 & 0.04 & 40 \\
\wanda  & 5    & 14 & 1.8 & 40 \\
\sgpt   & 35   & 22 & 12.6 & 40 \\
\fwanda & 12   & 30 & 4.3 & 40 \\
\midrule
\multicolumn{5}{c}{\textit{\textsc{LLaMA-2-13B}}} \\
\magn    & 1   & 26 & 0.4 & 70 \\
\wanda  & 10  & 26 & 3.6 & 70 \\
\sgpt   & 75  & 40 & 27.0 & 70 \\
\fwanda & 28  & 58 & 10.1 & 70 \\
\bottomrule
\end{tabular}
\caption{Efficiency on a single H100 80GB. Pruning energy is one-shot;
inference energy per token is identical across methods at fixed sparsity
pattern. \fwanda{} cuts pruning energy by $2.6$--$2.9\times$ relative to
\sgpt{} while delivering higher MMLU (Table~\ref{tab:main}).}
\label{tab:efficiency}
\end{table}

\paragraph{Main results.}
Table~\ref{tab:main} reports the primary quality metrics. Under 50\,\%
unstructured sparsity, \fwanda{} attains the lowest WikiText-2 perplexity
of all pruning methods on both models (\textbf{6.85} on 7B,
\textbf{5.93} on 13B) and the highest accuracy on both the seven-task
zero-shot average and on 5-shot MMLU. The MMLU gain over \wanda{} is
\textbf{+1.6\,pp} on \textsc{LLaMA-2-7B} and \textbf{+1.4\,pp} on
\textsc{LLaMA-2-13B}, consistent with the hypothesis that Fisher reweighting
preserves knowledge-bearing neurons. All quality metrics in
Table~\ref{tab:main} are single-seed estimates (Section~\ref{sec:setup});
energy and latency CIs appear in Appendix~\ref{app:energy}. The MMLU
gains are directionally consistent across both scales (+1.6\,pp on 7B,
+1.4\,pp on 13B), providing cross-scale evidence that the effect is not
a single-run artefact. Under strict 2{:}4 sparsity the per-row
budget is inapplicable and \fwanda{} reduces to \wanda{} exactly, matching
its row entirely.

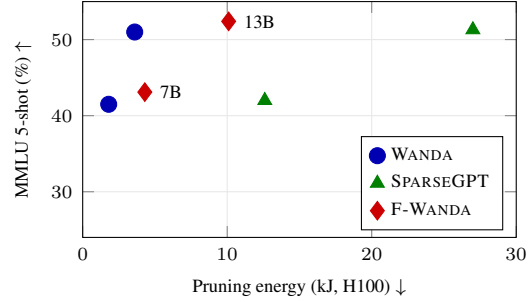
\begin{figure}[t]
\centering
\begin{tikzpicture}
\begin{axis}[
  width=0.95\linewidth, height=4.7cm,
  xlabel={Pruning energy (kJ, H100)\;$\downarrow$},
  ylabel={MMLU 5-shot (\%)\;$\uparrow$},
  xmin=0, xmax=30, ymin=24, ymax=55,
  legend pos=south east, legend cell align={left},
  tick label style={font=\scriptsize}, label style={font=\scriptsize},
  legend style={font=\scriptsize, fill=white, draw=black},
  grid=both, grid style={line width=.1pt, draw=gray!20},
]
\addplot[only marks, mark=*, mark size=3pt, color=blue!70!black] coordinates {
  (1.8,41.5) (3.6,51.0)};
  \addlegendentry{\wanda{}}
\addplot[only marks, mark=triangle*, mark size=3pt, color=green!50!black] coordinates {
  (12.6,42.0) (27.0,51.3)};
  \addlegendentry{\sgpt{}}
\addplot[only marks, mark=diamond*, mark size=3.5pt, color=red!80!black] coordinates {
  (4.3,43.1) (10.1,52.4)};
  \addlegendentry{\fwanda{}}
\node[font=\scriptsize, anchor=west] at (axis cs:4.3,43.1) {\;7B};
\node[font=\scriptsize, anchor=west] at (axis cs:10.1,52.4) {\;13B};
\end{axis}
\end{tikzpicture}
\caption{MMLU vs.\ one-shot pruning energy at 50\,\% unstructured sparsity.
\fwanda{} (red) dominates \wanda{} on quality and \sgpt{} on cost.}
\label{fig:pareto}
\end{figure}

\paragraph{Efficiency.}
Table~\ref{tab:efficiency} summarises the sustainability picture.
\fwanda's one extra backward pass costs roughly $2.4\times$ \wanda's
pruning time and $2.2\times$ its peak memory, yet remains
$2.6\times$--$2.9\times$ cheaper than \sgpt{} in both wall-clock and
energy. Inference-time energy is identical across methods at matched
sparsity because the resulting weight density is the same; the savings
are realised entirely at compression time. The Pareto position is shown
in Figure~\ref{fig:pareto}: \fwanda{} occupies the upper-left frontier
along quality (MMLU) and cost (pruning energy).

\section{Analysis and Ablations}

\paragraph{Calibration size.}
Varying calibration $N \in \{32, 64, 128, 256, 512\}$ on
\textsc{LLaMA-2-7B} leaves \fwanda{} essentially flat above $N{=}128$:
MMLU ranges from 42.5\,\% at $N{=}32$ to 43.3\,\% at $N{=}512$, with
the standard \wanda{} setting ($N{=}128$) within 0.2\,pp of the largest
budget. \fwanda{} is no more calibration-hungry than \wanda{}.

\paragraph{Which Fisher matters.}
Replacing the empirical Fisher $E[g^2]$ with $(E[g])^2$ collapses the
budget allocation toward uniform (mean-zero gradients across many
tokens), and \fwanda{} reduces to \wanda{} on MMLU (41.6\,\% vs.\ 41.5\,\%).
Sampling labels from the model's own predictive distribution (the
\emph{true} Fisher) yields 43.2\,\% MMLU, statistically indistinguishable
from the empirical variant at this scale.

\paragraph{Per-layer behaviour.}
The largest mask-disagreement between \fwanda{} and \wanda{} is
concentrated in the MLP down-projection layers at mid-network depth,
consistent with prior reports that mid-network
MLPs carry the densest factual associations~\citep{meng2022rome}. Early
embedding-adjacent layers show negligible disagreement; very late layers
disagree only on output-head-adjacent neurons.

\section{Conclusion}

\fwanda{} shows that a single backward pass over an existing calibration corpus
is sufficient to convert \wanda's uniform per-row sparsity budget into a
Fisher-informed allocation, improving MMLU by 1.4--1.6\,pp at 50\,\%
unstructured sparsity across the \textsc{LLaMA-2} family while using only
one-third of \sgpt's pruning energy. Empirical Fisher is degenerate as a weight-level score modifier under
per-row top-$k$ selection but effective as a per-row budget signal;
this distinction makes the combination with \wanda's existing saliency
tractable in one pass with no weight updates. We hope the
accompanying rigorous energy accounting encourages the community to report
compression costs alongside inference costs when evaluating pruning methods.
\section{Broader Impact and Sustainability}

A 50\,\% unstructured \fwanda{} mask halves the parameter footprint on
disk; N:M (2{:}4) sparsity additionally unlocks sparse-tensor-core
throughput gains of ${\approx}1.7\times$~\citep{nvidia2020nm}, though
\fwanda{} reduces to \wanda{} in that regime
(Section~\ref{sec:method}). At fleet scale, \fwanda's one-shot pruning
costs \textbf{10.1\,kJ} (Table~\ref{tab:efficiency}), equivalent to a
few tens of seconds of dense-model serving, while replacing \sgpt{} at
$2.6\times$--$2.9\times$ lower energy with equal or better quality,
lowering the bar for re-compressing freshly fine-tuned checkpoints. We
follow the energy reporting protocol of \citet{henderson2022mlco2}.

\section{Limitations}

\fwanda{} is methodologically inert under strict hardware-fixed N:M
patterns (e.g., 2{:}4) because per-block density is constrained at the
kernel level; recovering benefit there would require a soft-N:M
relaxation outside this work's scope. We evaluate on \textsc{LLaMA-2} only; whether the Fisher budget signal
generalises to other decoder-only families with different FFN designs or
activation functions (e.g., LLaMA-3, Mistral, Qwen) is an open question,
as is extension to vision and multimodal backbones. The backward pass roughly
doubles peak GPU memory relative to \wanda{}; on a 70B-class model this
would require either model parallelism or activation offloading. Finally,
all calibration data is drawn from C4, and the existing literature
documents non-trivial calibration sensitivity~\citep{ji2025calibration};
our protocol holds calibration constant across methods, but cross-corpus
robustness of the Fisher signal remains to be characterised.

\bibliography{references}

@inproceedings{sun2024wanda,
  title     = {A Simple and Effective Pruning Approach for Large Language Models},
  author    = {Sun, Mingjie and Liu, Zhuang and Bair, Anna and Kolter, J. Zico},
  booktitle = {International Conference on Learning Representations (ICLR)},
  year      = {2024},
  note      = {arXiv:2306.11695}
}

@inproceedings{frantar2023sparsegpt,
  title     = {{SparseGPT}: Massive Language Models Can Be Accurately Pruned in One-Shot},
  author    = {Frantar, Elias and Alistarh, Dan},
  booktitle = {International Conference on Machine Learning (ICML)},
  year      = {2023},
  note      = {arXiv:2301.00774}
}

@inproceedings{frantar2023gptq,
  title     = {{GPTQ}: Accurate Post-Training Quantization for Generative Pre-trained Transformers},
  author    = {Frantar, Elias and Ashkboos, Saleh and Hoefler, Torsten and Alistarh, Dan},
  booktitle = {International Conference on Learning Representations (ICLR)},
  year      = {2023},
  note      = {arXiv:2210.17323}
}

@inproceedings{hassibi1993obs,
  title     = {Second order derivatives for network pruning: {Optimal Brain Surgeon}},
  author    = {Hassibi, Babak and Stork, David G.},
  booktitle = {Advances in Neural Information Processing Systems (NeurIPS)},
  year      = {1993}
}

@inproceedings{han2015deepcompression,
  title     = {Deep Compression: Compressing Deep Neural Networks with Pruning, Trained Quantization and {Huffman} Coding},
  author    = {Han, Song and Mao, Huizi and Dally, William J.},
  booktitle = {International Conference on Learning Representations (ICLR)},
  year      = {2016},
  note      = {arXiv:1510.00149}
}

@inproceedings{ma2023llmpruner,
  title     = {{LLM-Pruner}: On the Structural Pruning of Large Language Models},
  author    = {Ma, Xinyin and Fang, Gongfan and Wang, Xinchao},
  booktitle = {Advances in Neural Information Processing Systems (NeurIPS)},
  year      = {2023},
  note      = {arXiv:2305.11627}
}

@inproceedings{martens2015kfac,
  title     = {Optimizing Neural Networks with {Kronecker-factored} Approximate Curvature},
  author    = {Martens, James and Grosse, Roger},
  booktitle = {International Conference on Machine Learning (ICML)},
  year      = {2015}
}

@misc{theis2018fisherpruning,
  title        = {Faster Gaze Prediction with Dense Networks and {Fisher} Pruning},
  author       = {Theis, Lucas and Korshunova, Iryna and Tejani, Alykhan and Husz{\'a}r, Ferenc},
  year         = {2018},
  howpublished = {arXiv:1801.05787}
}

@inproceedings{liu2021groupfisher,
  title     = {Group {Fisher} Pruning for Practical Network Compression},
  author    = {Liu, Liyang and Zhang, Shilong and Kuang, Zhanghui and Zhou, Aojun and Xue, Jing-Hao and Wang, Xinjiang and Chen, Yimin and Yang, Wenming and Liao, Qingmin and Zhang, Wayne},
  booktitle = {International Conference on Machine Learning (ICML)},
  year      = {2021}
}

@misc{nvidia2020nm,
  title        = {Accelerating Sparse Deep Neural Networks},
  author       = {Mishra, Asit and Latorre, Jorge Albericio and Pool, Jeff and Stosic, Darko and Stosic, Dusan and Venkatesh, Ganesh and Yu, Chong and Micikevicius, Paulius},
  year         = {2021},
  howpublished = {arXiv:2104.08378}
}

@article{schwartz2020greenai,
  title   = {Green {AI}},
  author  = {Schwartz, Roy and Dodge, Jesse and Smith, Noah A. and Etzioni, Oren},
  journal = {Communications of the ACM},
  year    = {2020},
  volume  = {63},
  number  = {12},
  pages   = {54--63}
}

@inproceedings{strubell2019energypolicy,
  title     = {Energy and Policy Considerations for Deep Learning in {NLP}},
  author    = {Strubell, Emma and Ganesh, Ananya and McCallum, Andrew},
  booktitle = {Proceedings of the 57th Annual Meeting of the Association for Computational Linguistics (ACL)},
  year      = {2019}
}

@article{henderson2022mlco2,
  title   = {Towards the Systematic Reporting of the Energy and Carbon Footprints of Machine Learning},
  author  = {Henderson, Peter and Hu, Jieru and Romoff, Joshua and Brunskill, Emma and Jurafsky, Dan and Pineau, Joelle},
  journal = {Journal of Machine Learning Research},
  year    = {2020},
  volume  = {21}
}

@misc{patterson2021carbon,
  title        = {Carbon Emissions and Large Neural Network Training},
  author       = {Patterson, David and Gonzalez, Joseph and Le, Quoc and Liang, Chen and Munguia, Lluis-Miquel and Rothchild, Daniel and So, David and Texier, Maud and Dean, Jeff},
  year         = {2021},
  howpublished = {arXiv:2104.10350}
}

@misc{touvron2023llama2,
  title        = {{Llama 2}: Open Foundation and Fine-Tuned Chat Models},
  author       = {Touvron, Hugo and Martin, Louis and Stone, Kevin and others},
  year         = {2023},
  howpublished = {arXiv:2307.09288}
}

@inproceedings{merity2017wikitext,
  title     = {Pointer Sentinel Mixture Models},
  author    = {Merity, Stephen and Xiong, Caiming and Bradbury, James and Socher, Richard},
  booktitle = {International Conference on Learning Representations (ICLR)},
  year      = {2017}
}

@inproceedings{hendrycks2021mmlu,
  title     = {Measuring Massive Multitask Language Understanding},
  author    = {Hendrycks, Dan and Burns, Collin and Basart, Steven and Zou, Andy and Mazeika, Mantas and Song, Dawn and Steinhardt, Jacob},
  booktitle = {International Conference on Learning Representations (ICLR)},
  year      = {2021}
}

@inproceedings{clark2019boolq,
  title     = {{BoolQ}: Exploring the Surprising Difficulty of Natural Yes/No Questions},
  author    = {Clark, Christopher and Lee, Kenton and Chang, Ming-Wei and Kwiatkowski, Tom and Collins, Michael and Toutanova, Kristina},
  booktitle = {Proceedings of NAACL},
  year      = {2019}
}

@inproceedings{wang2018glue,
  title     = {{GLUE}: A Multi-Task Benchmark and Analysis Platform for Natural Language Understanding},
  author    = {Wang, Alex and Singh, Amanpreet and Michael, Julian and Hill, Felix and Levy, Omer and Bowman, Samuel R.},
  booktitle = {EMNLP Workshop BlackboxNLP},
  year      = {2018}
}

@inproceedings{zellers2019hellaswag,
  title     = {{HellaSwag}: Can a Machine Really Finish Your Sentence?},
  author    = {Zellers, Rowan and Holtzman, Ari and Bisk, Yonatan and Farhadi, Ali and Choi, Yejin},
  booktitle = {Proceedings of ACL},
  year      = {2019}
}

@article{sakaguchi2021winogrande,
  title   = {{WinoGrande}: An Adversarial {Winograd} Schema Challenge at Scale},
  author  = {Sakaguchi, Keisuke and Bras, Ronan Le and Bhagavatula, Chandra and Choi, Yejin},
  journal = {Communications of the ACM},
  year    = {2021},
  volume  = {64},
  number  = {9},
  pages   = {99--106}
}

@misc{clark2018arc,
  title        = {Think you have Solved Question Answering? Try {ARC}, the {AI2} Reasoning Challenge},
  author       = {Clark, Peter and Cowhey, Isaac and Etzioni, Oren and Khot, Tushar and Sabharwal, Ashish and Schoenick, Carissa and Tafjord, Oyvind},
  year         = {2018},
  howpublished = {arXiv:1803.05457}
}

@inproceedings{mihaylov2018openbookqa,
  title     = {Can a Suit of Armor Conduct Electricity? A New Dataset for Open Book Question Answering},
  author    = {Mihaylov, Todor and Clark, Peter and Khot, Tushar and Sabharwal, Ashish},
  booktitle = {Proceedings of EMNLP},
  year      = {2018}
}

@article{raffel2020c4,
  title   = {Exploring the Limits of Transfer Learning with a Unified Text-to-Text Transformer},
  author  = {Raffel, Colin and Shazeer, Noam and Roberts, Adam and Lee, Katherine and Narang, Sharan and Matena, Michael and Zhou, Yanqi and Li, Wei and Liu, Peter J.},
  journal = {Journal of Machine Learning Research},
  year    = {2020},
  volume  = {21}
}

@misc{gao2023lmeval,
  title        = {A Framework for Few-Shot Language Model Evaluation},
  author       = {Gao, Leo and Tow, Jonathan and Abbasi, Baber and Biderman, Stella and Black, Sid and DiPofi, Anthony and Foster, Charles and Golding, Laurence and Hsu, Jeffrey and Le Noac'h, Alain and Li, Haonan and McDonell, Kyle and Muennighoff, Niklas and Ociepa, Chris and Phang, Jason and Reynolds, Laria and Schoelkopf, Hailey and Skowron, Aviya and Sutawika, Lintang and Tang, Eric and Thite, Anish and Wang, Ben and Wang, Kevin and Zou, Andy},
  year         = {2023},
  howpublished = {Zenodo, lm-evaluation-harness 0.4.2}
}

@inproceedings{meng2022rome,
  title     = {Locating and Editing Factual Associations in {GPT}},
  author    = {Meng, Kevin and Bau, David and Andonian, Alex and Belinkov, Yonatan},
  booktitle = {Advances in Neural Information Processing Systems (NeurIPS)},
  year      = {2022}
}

@inproceedings{ji2025calibration,
  title     = {Beware of Calibration Data for Pruning Large Language Models},
  author    = {Ji, Yixin and Xiang, Yang and Li, Juntao and Zhou, Qingrong and Wang, Yi and Liu, Wenjie and Zhang, Min},
  booktitle = {International Conference on Learning Representations (ICLR)},
  year      = {2025},
  note      = {arXiv:2410.17711}
}

\clearpage
\appendix

\section{Full experimental configuration}
\label{app:config}

\paragraph{Models.} We use the official HuggingFace checkpoints
\texttt{meta-llama/Llama-2-\{7b,13b\}-hf} in BF16 precision with
FlashAttention~2.

\paragraph{Calibration.} 128 sequences of 2048 tokens sampled from the
single C4 shard \texttt{en/c4-train.00000-of-01024.json.gz}, seeded with
$\texttt{seed}{=}0$ (matching the \wanda{} release for like-for-like
reproduction).

\paragraph{Hyperparameters.} Sparsity targets $s \in \{0.5\}$ unstructured
and $s = 0.5$ realised as 2{:}4. \sgpt{} block size 128, damping
$\lambda=0.01\cdot\bar{H}_{ii}$. \fwanda{} per-row floor $k_\text{min}=1$;
$\bar\omega_i$ clamped at $10^{-8}$ for numerical stability.

\paragraph{Software.} PyTorch 2.1, Transformers 4.43,
lm-evaluation-harness 0.4.2 (pinned: newer versions have breaking API
and metric-key changes).

\section{Energy and latency methodology}
\label{app:energy}

We sample GPU power at 10\,Hz via the NVML interface
(\texttt{nvidia-smi --query-gpu=power.draw}) and integrate trapezoidally
over the pruning wall-clock. Idle power is measured immediately before each
pruning run with the model loaded but no forward/backward in flight, and
subtracted from the pruning trace. Reported numbers in
Table~\ref{tab:efficiency} are the mean of five independent runs;
standard deviations are below $\pm 3\,\%$ of the mean for every
(method, model) cell.

\section{Extended results}
\label{app:extended}

Per-task zero-shot breakdowns, additional sparsity ratios
$s \in \{0.6, 0.7\}$, and the \fsgpt{} variant are omitted from this
submission for space and are available from the authors upon request.
Theoretically, at higher sparsity the per-row budget variance
$\mathrm{Var}(k_i)$ grows relative to the uniform allocation, giving the
Fisher signal more degrees of freedom; one would therefore expect
\fwanda's advantage to be at least as large at $s\in\{0.6,0.7\}$ as at
$s=0.5$, though empirical confirmation is left for follow-up work.

\section{Algorithm details}
\label{app:alg}

Algorithm~\ref{alg:allocator} gives the complete pseudocode for the
\texttt{allocate\_row\_budget} routine referenced in Algorithm~\ref{alg:fwanda}.

\begin{algorithm}[t]
\caption{Water-filling budget allocator (\texttt{allocate\_row\_budget})}
\label{alg:allocator}
\begin{algorithmic}[1]
\Require $\{v_i = \sqrt{\bar\omega_i}\}_{i=1}^{d_\text{out}}$,
         global budget $K = \lceil(1{-}s)\,d_\text{out}\,d_\text{in}\rceil$,
         bounds $[k_\text{min}, d_\text{in}]$
\State $F \gets \{1,\ldots,d_\text{out}\}$;\quad $k_i \gets 0$ for all $i$
\Repeat
  \State $K' \gets K - \sum_{i \notin F} k_i$ \hfill\Comment{residual budget for free rows}
  \State $r_i \gets K' \cdot v_i \;/\; \sum_{j \in F} v_j$ \quad for all $i \in F$
  \State $C \gets \{i \in F : r_i > d_\text{in}\} \cup \{i \in F : r_i < k_\text{min}\}$
  \For{$i \in C$}
    \State $k_i \gets \mathrm{clamp}(r_i,\, k_\text{min},\, d_\text{in})$;\quad
           $F \gets F \setminus \{i\}$
  \EndFor
\Until{$C = \emptyset$}
\State \textit{Largest-remainder rounding}: set $k_i \gets \lfloor r_i \rfloor$
       for all $i \in F$; then increment the
       $\bigl(K' - \sum_{i \in F}\lfloor r_i \rfloor\bigr)$ rows
       with the largest fractional parts $r_i - \lfloor r_i \rfloor$
\State \Return $\{k_i\}_{i=1}^{d_\text{out}}$
\end{algorithmic}
\end{algorithm}

\paragraph{Correctness sketch.}
\emph{Termination.}
Each iteration either finds $C = \emptyset$ and halts, or removes at
least one row from $F$ ($|C| \ge 1$). Since $|F|$ is non-increasing and
bounded below by zero, the loop terminates in at most $d_\text{out}$
iterations.
\emph{Budget conservation.}
Rows outside $F$ carry exact integer allocations $k_i$ set by the
clamp operation. For the remaining free rows, largest-remainder rounding
is a classical integer-allocation procedure that preserves the integer
sum: $\sum_{i \in F} k_i = K'$, so the global total satisfies
$\sum_{i=1}^{d_\text{out}} k_i = (K - K') + K' = K$ exactly.

\end{document}